\documentclass[conference]{IEEEtran}

\usepackage{cite}
\usepackage{amsmath}
\usepackage{algorithmic}
\usepackage{array}
\usepackage{url}
\usepackage{color}
\usepackage{soul}
\usepackage{enumerate}

\ifCLASSINFOpdf
  \usepackage[pdftex]{graphicx}
  \graphicspath{{../pdf/}{../jpeg/}}
\else
  \usepackage[dvips]{graphicx}
  \graphicspath{{../eps/}}
  \DeclareGraphicsExtensions{.eps}
\fi

\ifCLASSOPTIONcompsoc
  \usepackage[caption=false,font=normalsize,labelfont=sf,textfont=sf]{subfig}
\else
  \usepackage[caption=false,font=footnotesize]{subfig}
\fi

\ifCLASSOPTIONcaptionsoff
  \usepackage[nomarkers]{endfloat}
 \let\MYoriglatexcaption\caption
 \renewcommand{\caption}[2][\relax]{\MYoriglatexcaption[#2]{#2}}
\fi

\begin{document}

\title{STATE--OF--THE--ART AND GAPS FOR DEEP LEARNING ON LIMITED TRAINING DATA IN REMOTE SENSING}

%

\author{\IEEEauthorblockN{John E. Ball$^{1}$,
Derek T. Anderson $^{2}$, and
Pan Wei$^{1}$}
\IEEEauthorblockA{$^{1}$Department of Electrical and Computer Engineering, Mississippi State University\\
$^{2}$Department of Electrical Engineering and Computer Science, The University of Missouri
}}

\maketitle

%
%
\begin{abstract}

Deep learning usually requires big data, with respect to both volume and variety. However, most remote sensing applications only have limited training data, of which a small subset is labeled. Herein, we review three state-of-the-art approaches in deep learning to combat this challenge. The first topic is transfer learning, in which some aspects of one domain, e.g., features, are transferred to another domain. The next is unsupervised learning, e.g., autoencoders, which operate on unlabeled data. The last is generative adversarial networks, which can generate realistic looking data that can fool the likes of both a deep learning network and human. The aim of this article is to raise awareness of this dilemma, to direct the reader to existing work and to highlight current gaps that need solving. 

\end{abstract}


\IEEEpeerreviewmaketitle

%
%
\section{Introduction}
\label{sec:Introduction}

\noindent Deep learning (DL) has gained much attention in the research communities, due in part to its significant performance leap in comparison to traditional hand-crafted (often called shallow) solutions. Deep networks (DNs) can learn extremely complicated hierarchical features and decision boundaries from the training data. DNs are typically composed of many consecutive layers, while standard shallow neural networks are usually composed of only a few layers.

DNs can have from hundreds to millions of parameters that must be learned. Traditional DNs are trained using large datasets of imagery. However, the datasets available to remote sensing researchers are typically very limited, and often the number of labeled samples are usually also limited.

DL is significantly impacting areas of research, including computer vision, image processing, and remote sensing. In \cite{ball2017comprehensive}, we cite nine major challenges facing DL in remote sensing. This paper examines one of these challenges, namely the lack of training data (which is a typical problem in remote sensing data analysis). The community has started to address these challenges using several different approaches; 1) transfer learning, 2) unsupervised learning, and 3) generative adversarial networks (GANs). This is still an active area of research and the ultimate solutions (if one exists) is still an open question in the remote sensing field.

%
%
\subsection{Approach 1: Transfer Learning}

\noindent A well-known concern about machine learning in general is over fitting. Generalizability is our ultimate goal, yet there is no guarantee that our solutions will move from the training (source) domain to a new (target) domain. According to Tuia et al. \cite{tuia2016domain} and Pan et al. \cite{pan2010survey}, transfer learning seeks to learn from one area to another. The process of successful transfer learning is still an open issue in remote sensing. Moreover, transfer learning is also an open question in the general field of deep learning \cite{Zhang2016, ball2017comprehensive}. Pan et al. \cite{pan2010survey} point out that typically in remote sensing applications, when changing sensors or changing to a different part of a large image or other imagery collected at different times, the transfer fails. Also, transfer between images where the number and types of endmembers are different has very few studies.

Although in general these open questions remain, we do note that many researchers have found clever methods to mitigate limited training data. 

Ghazi et al. \cite{ghazi2017plant} suggest that two options for transfer learning are to utilize a pre-trained network and learn new features in the imagery to be analyzed, or fine-tune the weights of the pre-trained network using the imagery to be analyzed. The choice depends on the size and similarity of the training and testing datasets.

Othman et al. \cite{Othman2016Using} utilized transfer learning by training on the ILSVRC-12 challenge dataset, which has 1.2 million $224 \times 224$ RGB images belonging to 1,000 classes. The trained system was applied to the UC Merced Land Use \cite{yang2010bag} and Banja-Luka \cite{risojevic2011gabor} datasets. Iftene et al. \cite{iftene2016very} applied a pretrained CaffeNet and GoogleNet models on the ImageNet dataset, and then applied the results to the VHR imagery denoted the WHU-RS dataset \cite{chatfield2014return, sheng2012high}. In \cite{CGINets}, Scott et al. used data augmentation, they replaced and retrained the classification layers of GoogleNet, CaffeNet and ResNet, and they also updated the feature weights (e.g., convolution layers). In \cite{scott2017fusion}, they fused these transfer learned deep nets using the Choquet integral on the UC Merced and  data sets.

In \cite{xie2015transfer}, Xie et al. circumvented their small amount of training data for satellite imagery and poverty analysis by training a CNN on night-time light-intensity imagery, which was used as a data proxy for poverty analysis. The network is able to learn filters identifying roads, buildings and farmlands simply from the night-time data. In their system, ImageNet \cite{russakovsky2015imagenet}, which has over 14 million images and over 1,000 classes, was utilized. The transfer learning occurred from the ImageNet domain to the domain of night-time lights, and then the night-time lights features were transferred to a third domain, poverty mapping. Note that the poverty mapping domain had very limited training data available. This method demonstrates that multiple domains can be utilized for transfer learning.

Ghazi et al. \cite{ghazi2017plant} and Lee et al. \cite{Lee20171} used a pre-trained networks AlexNet, GoogLeNet and VGGNet on the LifeCLEF 2015 plant task dataset \cite{joly2016lifeclef} and MalayaKew dataset \cite{LeeCWR15} for plant identification. In \cite{Lee20171}, the authors discover that (1) leaf venation features are the best representative features, (2) multi-level representation in leaf data, demonstrating the hierarchical transformation of features from lower-level to higher-level abstraction, corresponding to species classes, and (3) these findings fit the hierarchical botanical definitions of leaf characteristics \cite{cope2012plant}.

Yang et al. \cite{Yang2016Two} utilized dual CNNs and transfer learning. In this method, the lower and middle layers can be trained on other scenes, and the top layers are trained on limited training samples. The two CNN outputs are concatenated and provide inputs to a fully connected layer for classification. 

Ding et al. \cite{ding2016deep} utilized transfer learning for automatic target recognition from mid-wave infrared to longwave IR. In their method, deep structures are designed to capture the information within source and target domains in a layer-wise fashion. In this way, more discriminative features across two domains could be extracted and more knowledge from source domain could be transferred. Secondly, a weighted class-wise adaptation scheme is proposed to couple the deep structures so that the target sample features become close to the same-class source sample features, and this helps to align the conditional and marginal distributions of both domains. Finally, the classifier is jointly learned with the deep structure. The task-driven scheme can feedback its classification error to refine the deep structures, and the discriminative deep features can also train a more effective classifier.

In summary, most remote sensing-based transfer learning work is focused on updating the weights (feature and/or classifier) of a DL solution from another context (the source domain) to the current task (target domain) based on available training data. However, numerous big questions remain. For example, i) does the current state-of-the-art truly apply to high spectral dimensionality HS imaging (HSI), ii) how does transfer learning work for HSI when the number and type of endmembers change, iii) how can transfer learning be made robust to imagery collected at different times and under different atmospheric conditions, iv) what happens if there is little commonality between the source and target domains, and v) how does transfer learning work in the context of multi-sensor fusion for remote sensing? 


%
%
\subsection{Approach 2: Unsupervised Training}

\noindent Whereas the last section focused on migrating a quality DL solution, here we address the reality that orders of magnitude more unlabeled versus labeled data exists in remote sensing. As such, its natural to ask how can we make use of this data in DL. Solutions put forth for DL include 1) autoencoders (AEs) and 2) dimensionality reduction techniques. 

AEs are usually unsupervised deep learning systems, where internal (latent) features are learned that can reproduce the input (or reproduce it with noise reduction), but the AE does not learn the trivial identity mapping. The AE can be linear, or by adding nonlinear elements (such as a \textit{tanh} layer), non-linear features can be generated. Moreover, the AE network can have a diabolo shape (taller layers on the ends and progressively smaller layers towards the middle of the network) -- this architecture forces the features to learn a smaller dimensional representation of the data. Unsupervised training means that the DL system learns from the data, but the data is unlabeled. Two main areas of research in unsupervised learning are noise reduction (denoising autoencoders - DAEs) and unsupervised feature extraction AEs. 

Petersson et al. \cite{petersson2016hyperspectral} suggested using sparse AEs (SAEs) to handle small training samples in HSI processing. Ma et al. \cite{ma2016spectral} put forth a DAE and used a collaborative representation-based classification, where each test sample can be linearly represented by the training samples in the same class with the minimum residual. In classification, features of each sample are approximated with a linear combination of features of all training samples within each class, and the label can be derived according to the class that best approximates the test features. Interested readers please see references $46$--$48$ in \cite{ma2016spectral} for more information on collaborative representations.

Tao et al. \cite{Tao2015Unsupervised} utilized a Stacked Sparse AE (SSAE) that was shown to be very generalizable and performed well in cases when there were limited training samples. The SSAE generates latent features that can be utilized to recreate a faithful reproduction of the input. The next layer takes the latent features from the first layer and again learns another latent representation. This process is repeated for each successive layer to create a hierarchy of features. Their experiments indicate that the learned spectral--spatial features are more discriminative for HSI classification compared with previously hand--engineered spectral--spatial features, especially when the training data are limited, and that the learned features appear not to be specific to a particular image but general in that they are applicable to multiple related images (e.g., images acquired by the same sensor but varying with location or time). This method then shows promise for both low numbers of training data and data with location/temporal variations.

Guo et al. \cite{guo2015hyperspectral} utilized a cascade of AEs. The first AE in this cascade is a DAE to denoise the hyperspectral (HS) data and a second AE with sparsity and non-negativity constraints to unmix the pixels and estimate the endmember signatures. The DAE utilized is the marginalized DAE put forth by Chen et al. \cite{chen2012marginalized}, which has a closed form solution (no backpropagation is required). There are still challenges to utilizing this DAE in large dimensional data (e.g. HS). The proposed network also hard codes the number of endmembers, which somewhat limits its applicability.

Zhou and Du \cite{Zhao2016Learning} proposed a deep learning spectral--spatial feature based classification framework that jointly uses dimensionality reduction and deep learning techniques for spectral and spatial feature extraction. A balanced local discriminant embedding algorithm (which balances the locality--preserving scatter matrix and between--class scatter matrix) is put forth for spectral feature extraction from high--dimensional HS datasets. This approach is similar in concept to Fisher's linear discriminant analysis, where the between--class margins are maximized and the within-class distances are minimized \cite{Fisher1936LDA}. A CNN is utilized to automatically find spatial--related features at high levels. The spectral and spatial features are then stacked for final processing.

In summary, a number of works have appeared in remote sensing for deep learning based on the application and extension of AEs, DAEs, and sparse AE architectures, to name a few. However, open challenges include: i) what architecture is optimal for transfer learning, ii) if different solutions exist, when should we pick one over the other, iii) what architectures can be used for high--dimensional (e.g., HSI) data, iv) how does this apply to multi--sensor fusion, and v) are there other ``architectures'' (algorithms) that can be effectively utilized?


%
%
\subsection{Approach 3: Generative Adversarial Networks (GANs)}

\noindent One of the sad and fundamental realities of sensor systems is that we will likely never have enough data or the cost of collecting and preparing the data might not be realistic. As such, it is natural to ask, can we create a system to create data? In the last few years, generative adversarial networks (GANs) were put forth to this end \cite{goodfellow2014generative}. GANs typically involve one part generator--who is learning the structure of the data and creates new data--and the discriminator--who is more--or--less the detector. These two networks, with various other little parts here and there in the community, work together by competing to improve each other. The networks are trained together, and the results (at least in the computer vision applications) can create near photo-realistic imagery that the classifier DL network cannot identify as fake \cite{goodfellow2014generative}. 


There has been a few papers recently utilizing GANs in remote sensing. A few examples are given herein. Costea et al. \cite{costea2017creating} utilized a dual--discriminator GANs to segment roadmaps from Aerial imagery. Enmoto et al. put forth a DL system for cloud removal from visible RGB satellite imagery by extending CGANs in ref. \cite{isola2016image} from RGB to multispectral imagery. Gong et al. in \cite{gong2017generative} used GANs for change detection.



Several papers have addressed HS classification using GANs. He et al. \cite{He2017GAN} tackled the problem of limited training data classification of HS data using GANs. Their proposed method is semi--supervised, which can make full use of the limited labeled samples as well as the larger number of unlabeled samples. The HS data is processed by a three-dimensional bilateral filter to extract spatial--spectral features. GANs are then trained on the spectral-spatial features. The  semi--supervised learning is achieved by adding samples from the generator to the features and increasing the dimension of the classifier output. The proposed method is effective, especially with a small number of training samples. 

Zhan et al. \cite{zhan2018semisupervised} used a 1D GAN and 1D CNN discriminator on the spectral data. The Indian Pines dataset was used. The GAN outperformed other methods including a CNN. This method did not utilize spatial information. Zhu et al. \cite{Zhu2018Generative} used both a 1D GAN (spectral only) and a more robust 3D (spectral/spatial) GAN. The GANs were conditional GANs, with class label inputs. 10 principal components were utilized for dimensionality reduction. Salias, Kennedy Space Center and Indian Pines were used for testing. Overall accuracies ranged from mid 80 \% to mid 90 \%. They found the GAN acts as a regularized to the discriminative CNN to help mitigate overfitting. Zhong et al. \cite{zhong2018generative} combined GANs with PGMs (Probabilistic Graphical Models) for HSI classification. The PGMs allow unsupervised data to be utilized. However, the overall accuracy numbers were in the mid 80 \% levels on Indian Pines and Pavia University datasets.

In summary, GANs have been used to improve processing of aerial imagery, change detection, and tackling limited training data in a spatial-spectral context. GANs have great potential, but they still must be trained (and remember they are dual deep networks), so limited training data is a very challenging problem as well as tuning. We note that in general, CNNs can be time--consuming to tune, and GANs can be notoriously difficult to tune properly (both the architectures, the training parameter, and the number of time the generator is run versus the discriminator). Solutions using GAN architectures viz. semi--supervised or non-supervised methods could offer significant improvements for remote sensing DL systems. A few open challenges include; i) GANs for high--dimensional (e.g., HSI) data, ii) GANs on low volume and/or variety training data, iii) and GANs for multi-sensor fusion--where the generator has to produce data for each sensor but now also take into account the complicated relationships between sensors.

%
%

%
%
\section{Conclusions}
\label{sec:Conclusion}

\noindent In this article, we examined transfer learning, unsupervised training and GANs to help mitigate the realistic and extreme challenge of limited training samples when applying deep learning in remote sensing applications. Namely, we reviewed existing state--of--the--art work and highlighted open challenges for the community. Common themes include how to address high--dimensional data (e.g., HSI), problems where the number and type of endmembers change in HSI, addressing data collected at different times and under different atmospheric conditions, what happens if there is little overlap between the source and target domain, and multi-sensor fusion. It is our belief that current work shows that there is indeed promise in these three technical topics. However, much future research is needed, which equates to opportunities for the community.

%

\ifCLASSOPTIONcaptionsoff
  \newpage
\fi

\bibliographystyle{IEEEtran}
\bibliography{refs.bib}

\end{document}